\documentclass[10pt,twocolumn,letterpaper]{article}

\usepackage[pagenumbers]{cvpr} 

\usepackage{cuted}
\usepackage{capt-of}
\usepackage{microtype}
\usepackage{makecell}
\usepackage{pifont}

\usepackage[dvipsnames]{xcolor}

\definecolor{cvprblue}{rgb}{0.21,0.49,0.74}
\usepackage[pagebackref,breaklinks,colorlinks,citecolor=cvprblue]{hyperref}

\def\paperID{XXXX} 
\def\confName{XXXX}
\def\confYear{XXXX}

\title{Jump Cut Smoothing for Talking Heads}

\author{
Xiaojuan Wang\textsuperscript{1}\space\space 
Taesung Park\textsuperscript{2}\space\space 
Yang Zhou\textsuperscript{2}\space\space 
Eli Shechtman\textsuperscript{2}\space\space 
Richard Zhang\textsuperscript{2}
\vspace{1mm} \\
\textsuperscript{1}University of Washington\qquad 
\textsuperscript{2}Adobe Research\vspace{1mm}\\
\href{https://jeanne-wang.github.io/jumpcutsmoothing/}{jeanne-wang.github.io/jumpcutsmoothing}
}
\begin{document}

\twocolumn[{%
\renewcommand\twocolumn[1][]{#1}%
\maketitle
\begin{center}
    \centering
   \includegraphics[width=.9\linewidth]{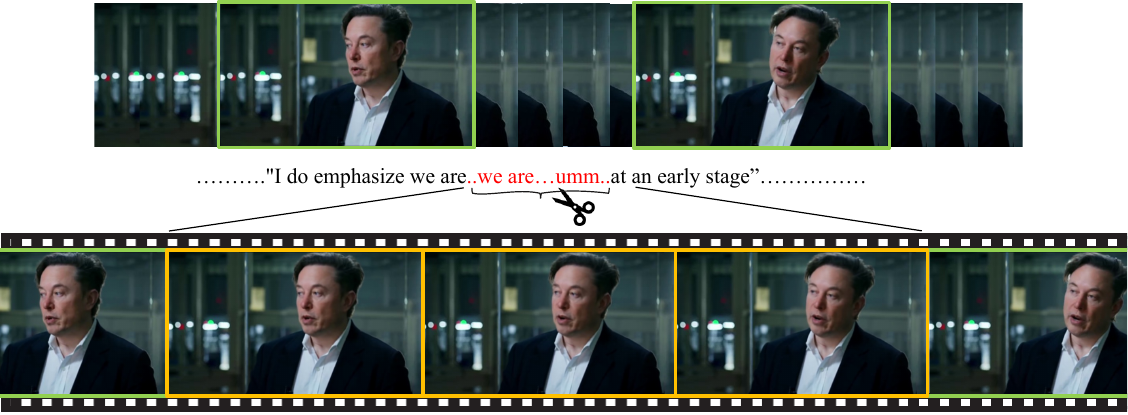} \\
    \captionof{figure}{{\bf Jump cut smoothing for filler words removal.} Given a talking head video, we remove the filler words and repetitive words (text in red color), and create a seamless transition for the jump cut  as shown in the second row.
\label{fig:teaser}}
\end{center}%
}]

\begin{abstract}
    A jump cut offers an abrupt, sometimes unwanted change in the viewing experience. We present a novel framework for smoothing these jump cuts, in the context of talking head videos. We leverage the appearance of the subject from the other source frames in the video, fusing it with a mid-level representation driven by DensePose keypoints and face landmarks. To achieve motion, we interpolate the keypoints and landmarks between the end frames around the cut. We then use an image translation network from the keypoints and source frames, to synthesize pixels. Because keypoints can contain errors, we propose a cross-modal attention scheme to select and pick the most appropriate source amongst multiple options for each key point. By leveraging this mid-level representation, our method can achieve stronger results than a strong video interpolation baseline. We demonstrate our method on various jump cuts in the talking head videos, such as cutting filler words, pauses, and even random cuts. Our experiments show that we can achieve seamless transitions, even in the challenging cases where the talking head rotates or moves drastically in the jump cut.
\end{abstract}

\section{Introduction}
\begin{figure*}[t!]
        \centering
	\includegraphics[width=1.0\linewidth]{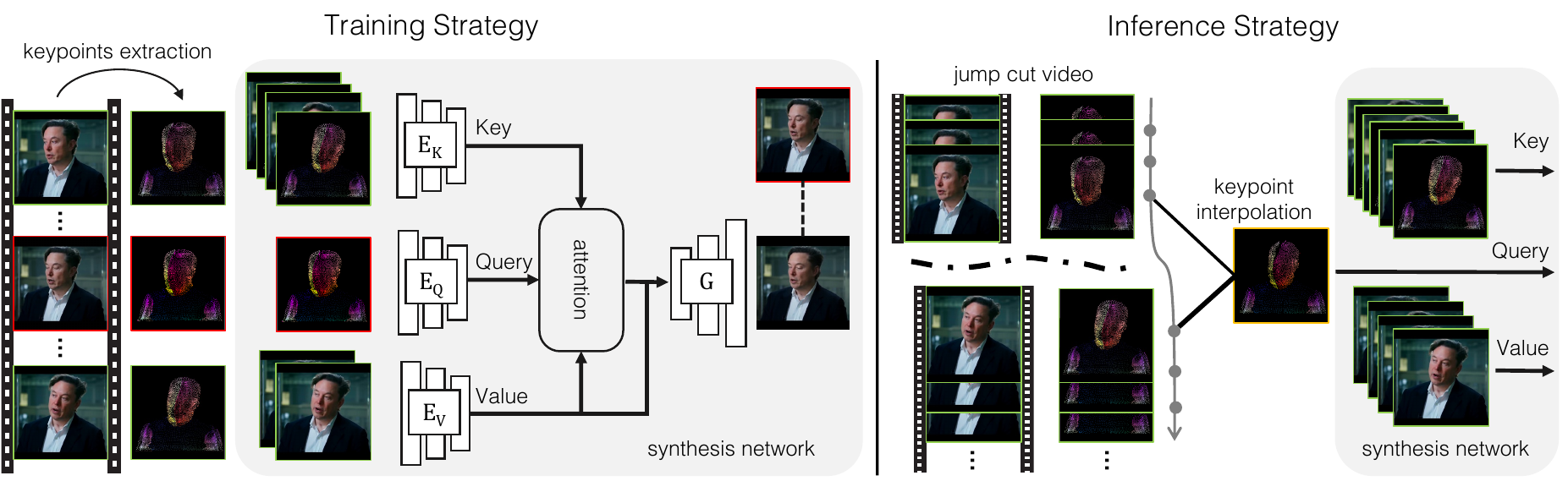}
	\caption{{\bf Method overview.} In the training stage, we randomly sample source (denoted in green rectangle) and target (denoted in red rectangle) frames, and extract their corresponding DensePose keypoints augmented with facial landmarks (not shown here for simplicity). Our method extracts source dense keypoint features as {\it key}, target dense keypoint feature as {\it query}, and source image features as {\it value}, then a cross attention is applied to get the values for the {\it query}, i.e., warped feature. This warped feature is fed into the generator inspired from Co-Mod GAN to synthesize a realistic target image compared with the ground truth target frame. For applying jump cut smoothing in the inference stage, we interpolate dense keypoints between jump cut end frames, and synthesize the transition frame with the interpolated keypoints (in yellow rectangle) sequence.}
	\label{fig:demo}
        \vspace{-0.3cm}
\end{figure*}
With the continuous advent of social media platforms, talking head videos have become increasingly popular.
Such videos usually focus on the head and upper body of a person who narrates an idea, story or concept while looking at the camera or an interviewer. Frequently, the subject may use filler words (``uhh''), stutter, make an unwanted pause, or repeat words. Directly removing these frames produces unnatural jump cuts, which can range from half a second to minutes long. In such time, large body movements, head movements, and hand gestures may occur. Can the video editor avoid presenting this unnatural experience to the viewer?

While some cuts can be smoothed through playing B-roll or kept intentionally for artistic choice, there lacks a robust tool for replacing the jump cut with a seamless transition. For example, Adobe's video editing tool, i.e., Premiere Pro includes a feature called {\it MorphCut}\footnote{https://helpx.adobe.com/premiere-pro/using/morph-cut.html} to help create a smooth transition. However, it fails when there are relatively large motion change such as head pose change or hand gesture change, resulting in noticeable motion blurs and other visual artifacts in the generated transitions. In influential work, Berthouzoz et al.~\cite{berthouzoz2012tools} generate hidden transitions by constructing a similarity graph between the raw frames and walking the graph. Such an approach is heavily limited by the contents of the video itself.
Compared to traditional methods~\cite{berthouzoz2012tools, kemelmacher2011exploring}, recent generative technology, offers the possibility of synthesizing new intermediate frames.

On the other hand, existing frame interpolation works utilize deep networks to perform video frame interpolation~\cite{niklaus2017video,reda2022film} have greatly advanced, showing the impressive ability to increase video temporal sampling, creating slow motion videos. However, these methods are not designed to handle relatively large or complex motion changes in the talking head videos, such as large head rotation and body translation.

In this paper, we propose to readdress the jump cut smoothing by synthesizing new intermediate frames guided by the interpolated motion. 
We leverage a mid-level motion representation, i.e., interpolated DensePose~\cite{guler2018densepose} keypoints between the cut end frames, augmented with face landmarks, and formulate our learning problem as transferring the appearances of multiple source images to the given target DensePose keypoints. This is related to image animation using single/multiple source images through the transfer of the motion of a driving video. However, these methods either cannot interpolate the motion learned from the driving video via the unsupervised way due to a lack of semantic correspondence, or they suffer from serious identity preservation through the jump cut end frames (see Fig.~\ref{fig:comp_to_image_animation}). During the learning process, simply warping the source appearances based on DensePose keypoints correspondence often loses details and produces unfaithful images because of misalignment of the DensePose, disocclusion, and so on. Besides, when there are multiple source images available, the naive way to average the warping results often causes blurry artifacts. Therefore we introduce a cross model attention mechanism  to improve the dense correspondence and help pick the appropriate source among multiple sources for each location in the target representation, which achieved better warping results to be used in the image generation network.

To conclude, we present a novel algorithm to smooth jump cuts in talking head videos through motion guided re-synthesis, a non-trivial task that requires balancing between realistic motion interpolation and preserving identity. For improved identity preservation, a larger informational bottleneck is needed, such as more keypoints or larger latent codes. Yet, more latent codes make motion interpolation challenging. We navigated this by (1) using more reliable and denser DensePose keypoints and face landmarks. (2) designing cross model attention, derived from initial DensePose keypoints correspondence, improving warping and allowing the synthesis network attend to more frames from the video and pick the most suitable features. (3) Employing smooth linear interpolation, as well as interpolation ablation augmentation, and training the model to handle missing correspondences between the jump cut end frames. Our experiments show that we can seamless bridge various jump cuts, even those with  significant head movements. 
\section{Related work}
\begin{figure*}[htbp!]
        \centering
	\includegraphics[width=1.\linewidth]{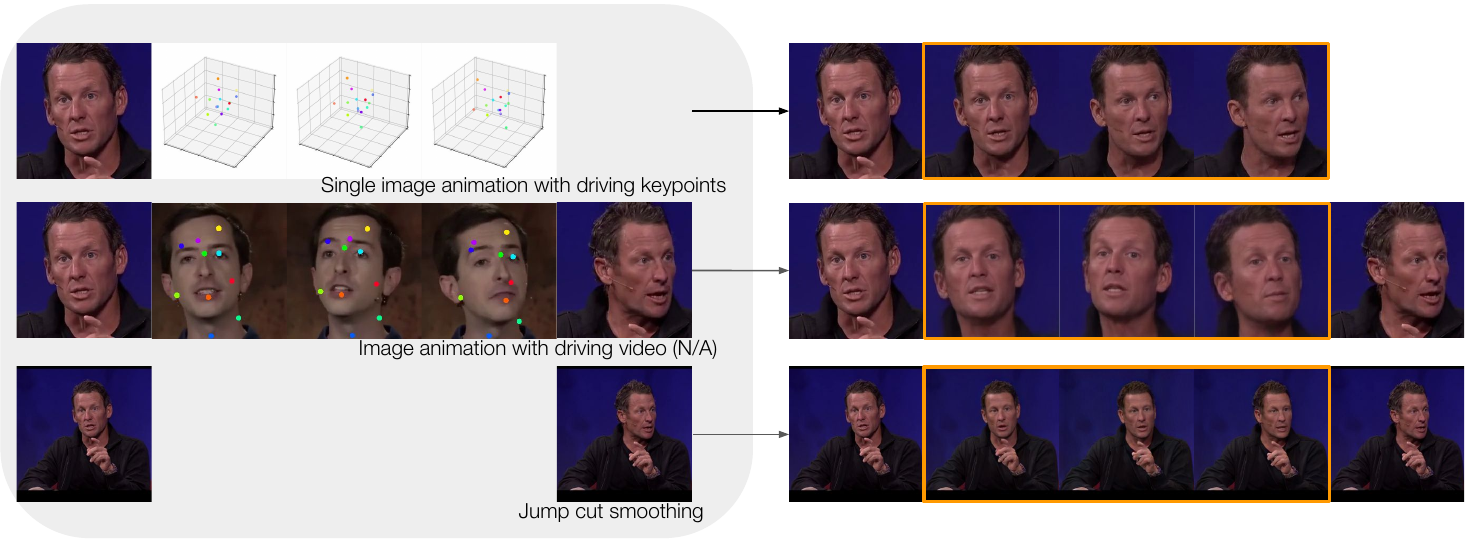}
	\caption{{\bf Image animation methods cannot be applied for jump cut smoothing.} Row\#1: Single image animation works (FaceVid2Vid~\cite{wang2021one}, Face2Face$^\rho$~\cite{yang2022face2face}) animate one of the cut end frames according to the key points sequence, neglecting the other end frame; Row\#2: Other works (FOMM~\cite{siarohin2019first}, ImplicitWarping~\cite{mallya2022implicit}) require a driving video for motion extraction, which is absent in our scenario. Row\#3: Our approach utilizes at least two cut end frames to generate the transition (shown in orange).}
	\label{fig:comp_to_image_animation}
 \vspace{-0.4cm}
\end{figure*}
{\noindent \bf CNN-based frame interpolation.} Frame interpolation, i.e., synthesizing intermediate images between a pair of input frames, is a long-standing research area in computer vision. Example applications include temporal up-sampling to increase refresh rate, create slow-motion videos and so on. Kernel-based methods~\cite{niklaus2017video} formulate frame interpolation as a convolution process, and they estimate spatially-adaptive convolution kernels for each output pixel and convolve the kernels with the input frames to generate a new frame. Flow-based methods~\cite{reda2022film, park2021asymmetric, niklaus2020softmax, lee2020adacof, park2020bmbc, huang2020rife, jiang2018super} first estimate optical flow between the input frames and then synthesize the middle images guided by the flow via either warping or splatting techniques. These methods have demonstrated impressive results for input frames with small motion change. However, in our talking head video editing situation where we cut out various lengths of filler words, unnecessary pauses, repeated words and so on, various motion change might happen in the jump cut such as large head rotation. This poses a major challenge for existing frame interpolation methods. Recently, ~\cite{reda2022film} proposed to use multi-scale feature extractor and present a “scale-agnostic” bidirectional flow estimation module to handle large motion between duplicate photos, and has achieved state-of-the-art results in frame interpolation. Even so, it cannot address large head rotation, translation, and complex motion change between the two frames. Different from these flow based methods, we tackle this problem with DensePose key point guided image synthesis, which gives flexibility and controlability over the synthesized intermediate images, and our framework allows us to utilize other additional frames in the video.

\vspace{0.1cm}
{\noindent \bf Image animation with driving video.} Our technique is related to the works ~\cite{mallya2022implicit, wang2021one, doukas2021headgan, zakharov2020fast, siarohin2019animating, siarohin2019first, siarohin2021motion, yang2022face2face, drobyshev2022megaportraits} in the image animation domain, which generates videos by animating a single source image of a subject using the motions of a driving video possibly containing a different subject. Typically, these techniques first extract motion representation from the driving video in an unsupervised way, and "warp" the source image feature to the learned motion for synthesis. However, popular methods like FOMM~\cite{siarohin2019first}, MegaPortraits~\cite{drobyshev2022megaportraits} and ImplicitWarping~\cite{mallya2022implicit}  often produce latent motions that lack semantic correspondence (see Row\#2 in Fig.~\ref{fig:comp_to_image_animation}), making them inapplicable for generating transition motions in our jump cut smoothing task. Even if the keypoints has semantic meaning, as seen in Face-Vid2Vid~\cite{wang2021one} which discerns 3D keypoints, head pose, and expression deformation, these methods grapple with identity preservation due to their reliance on a single image for animation (see Row\#1 in Fig.~\ref{fig:comp_to_image_animation}).
Recently ~\cite{mallya2022implicit} introduced Implicit Warping approach, which animates using multiple source images, leveraging cross-model attention to pick the most fitting keypoint features. Despite its impressive results, like FOMM~\cite{siarohin2019first}, it mandates a driving video to guide the motion. This stipulation, coupled with our lack of visual signals for the intermediate images we want to generate, renders it inapplicable for our purpose.

\vspace{0.1cm}
{\noindent \bf Attention in computer vision.} Given a query and a set of key-value pairs, the attention function outputs the value for the query, which is computed as a weighted sum of the values for the keys, and the weight is determined by  the similarity between query and the corresponding key. ~\cite{vaswani2017attention} proposed a transformer network stacked of attention layers and has achieved remarkable success in the task of machine translation. Since then, a number of  works extended such transformer networks to the compute vision domain for the task of image recognition~\cite{dosovitskiy2020image, liu2021swin, touvron2021training, wang2021pyramid}, image generation~\cite{jaegle2021perceiver, jaegle2021perceiver2, esser2021taming, hudson2021generative, parmar2018image} and achieved state-of-the-art numbers on the benchmark. There are also works using cross-modal attention where the queries, keys, and values are computed with different modalities such as vision and text~\cite{ye2019cross, xu2020cross}. More recently, diffusion models have gained significant success and roaring attention in high resolution image synthesis conditioning on various input modalities such as image to image and text to image. The image generator in the diffusion model uses UNet backbone with the cross-attention mechanism for the conditioning~\cite{rombach2022high, jaegle2021perceiver, jaegle2021perceiver2} and have proved great efficiency. In our method, we are doing cross attention between DensePose keypoint features of source and target, where they come from different poses. Upon this initial correspondence, our attention learns to  pick the most relevant source among the set of source images for each location in image. The warped feature are further fed into the generator network to output the intermediate frame. 
\section{Method}
We solve the jump cut smoothing problem in two stages (see Fig.~\ref{fig:demo}): in the training stage, given a set of source images, our network learns to generate a target image with the corresponding DensePose keypoints as motion guidance. In the inference stage, we synthesize each intermediate frame between the jump cut end frames guided by the interpolated DensePose keypoints and other available frames in the video. 
\subsection{DensePose keypoints representation}
We use DensePose~\cite{guler2018densepose} keypoints augmented with face landmarks as motion guidance to synthesize the corresponding image. Given an input image $I$ of size $(H\times W)$ and its continuous DensePose $P$, i.e., image-space UV coordinate map per body part, the DensePose keypoints $\{x_{k}\}_{k=1}^K$ are extracted from DensePose by quantizing the UV values. For each body part UV map, we discretize it into $n\times n$ cells with each cell representing a key point. We focus on the videos where only the upper body of a person appears, and thus there are $K = 14\times n\times n$ DensePose keypoints per image ($14$ is the number of upper body parts in then DensePose representation). The key point coordinates and UV value are the average of the pixel coordinates and UV values which fall into the corresponding cell. These keypoints with UVs are splatted into the grid of size $(H\times W)$ based on their coordinates, and then we get a discretized DensePose IUV. 
If there are not UV values in the DensePose fall into the cell, the corresponding key point is not visible, and will not be splatted.

\subsection{Cross model attention warping}
\begin{figure}[htbp]
    \centering
    \includegraphics[width=\linewidth]{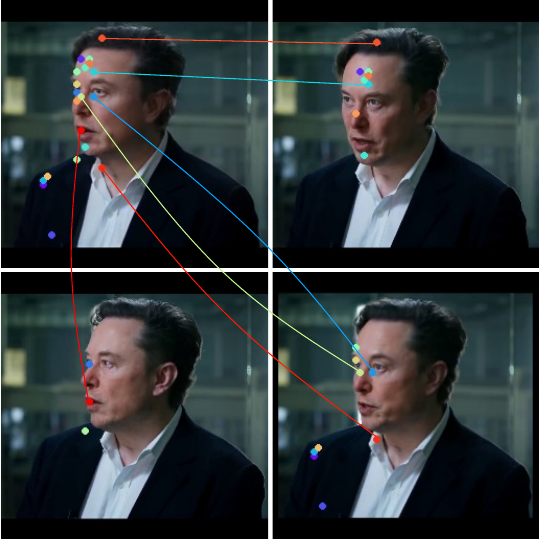}
    \caption{{\bf Visualization of learned correspondence with our attention mechanism.} The top left is our synthesized image given the other three images as sources. We highlight the locations in the synthesized image where the peak attention score $\ge 0.75$, and show their learned corresponding locations (marked with same color) in the source images. Our attention picks appropriate feature from different sources per location, e.g., for the blue point in the lower eyelid, our attention learned to associate with the eyelid feature in the bottom right source image. }
    \label{fig:attn_vis}
    \vspace{-18pt}
\end{figure}

Given a set of $N$ source images $\{I_i\}_{i=1}^N$ with their respective DensePose keypoints $\{x_{i, k}\}$ and the target DensePose keypoints $\{x_{t, k}\}$, we aim at generating a realistic target image by transferring the appearances from the source images. The basic idea is that we utilize this dense correspondence to warp the source image features to the target dense key point, then the warped feature is fed into a generator similar as Co-Mod GAN~\cite{zhao2021comodgan} to generate the respective realistic target image. Therefore it is critical to get a high quality warped feature. However, the DensePose based correspondence is often inaccurate, in addition, in the case of multiple source images, it is more natural to select the one that is closer to the target image instead of doing naive averaging. we propose to use attention mechanism to achieve this selection ability.

We adopt the commonly used scaled dot-product attention~\cite{vaswani2017attention}. The input of the attention function consists of queries and keys of dimension $d_k$, and values of dimension $d_v$. We compute the dot products of the query with all keys, divide each by $\sqrt{d_k}$ and apply a softmax function to obtain the weights on the values. The weight can be interpreted as the similarity of the query and the corresponding key, then the output per query is a weighted average of the values. In practice, the  set of queries are packed together into a matrix $Q$ of size $n_q\times d_k$. The keys and values are also packed together into matrices $K$ of size $n_k \times d_k$ and $V$ of size $n_k \times d_v$. The matrix of outputs of size $n_q\times d_v$ are computed as:
\begin{equation}
    \label{eq:attention}
    \text{Attention}(Q, K, V) = \text{Softmax}\bigg(\frac{Q K^T}{\sqrt{d_k}} \bigg) V
\end{equation}
$Q$ and $K$ are the feature representation of target and source DensePose keypoints respectively, and $V$ is the appearance feature of source images. We use StyleGAN2~\cite{karras2020analyzing} based encoder with a final project layer to map the feature to the dimension $d_k$ for query and key. When encoding the source DensePose keypoints, we also concatenate it with the source image. The source image appearance features are encoded by the encoder with similar structure. All of these encoders output outputs at $1/4$ resolution of the inputs. For example, the image and discretized DensePose keypoints input are of size $256\times 256$, the encoders produce $n_q = 64\times 64$ queries, and $n_k = 64\times64\times N$ keys for $N$ number of sources. The warped image features are computed according to Eq.~\ref{eq:attention}, and reshaped back to $64\times 64$, which will be fed into the generator to produce the according target image.

With cross model attention based warping, on the one hand, the feature representation of the DensePose keypoints is more robust and can correct the misalignment of the DensePose. On the other hand, our method can pick the most relevant source features among multiple source images per location (see Fig.~\ref{fig:attn_vis}). This is especially useful for jump cut smoothing where we have frames from the entire video as sources.

\subsection{Recursive synthesis for jump cut transitions} When creating the transition for jump cut smoothing, we firstly create a linearly interpolated dense keypoints sequence between the jump cut end frames, then each intermediate frame will be generated accordingly. This causes an issue when occlusion and disocclusion happen between the end frames. For example, when the speaker's head rotates from one side to the other side, part of the keypoints on the face disappears and another part of keypoints appears. We can only interpolate visible keypoints in both ends, which causes incomplete set of keypoints (see Fig.~\ref{fig:filmcomp}).  We simulate this case in the training by only use visible keypoints in all source images for the target, and make the generator to learn to inpaint the hole and generate realistic image.

\begin{figure}[htbp]
    \centering
    \includegraphics[width=\linewidth]{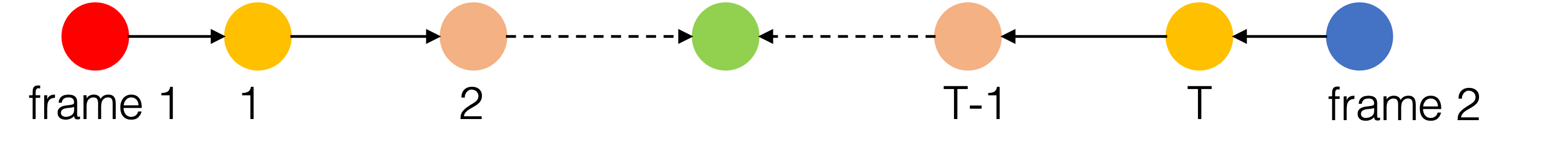} \\
    \caption{{\bf Recursive synthesis.} To fill in a jump cut with smooth, intermediate frames, we recursively fill in frames from the end towards the middle.}\label{fig:recursive}
    \vspace{-0.3cm}
\end{figure}

Since dense keypoints only model the foreground part, we additionally concatenating the two end frame features with the warped feature together as input to make the network learn to selectively copy the background from the end frames as well the remaining part that are not modeled by the DensePose such as hair. We further proposed a recursive synthesis, as shown in Fig.~\ref{fig:recursive}, for a transition sequence of length $T$, we start the synthesis from the two frames that are closest to the end frame, i.e., $I_1$ and $I_T$,  and then move towards the middle with the synthesized frames before as end frames to provide background information.

\subsection{Blended transition }\label{sec:blendedtransition}
Given a jump cut, we denote the frame before the cut as $I_m$ and the frame after the cut as $I_n$, and their respective DensePose keypoints as $\{x_{m,k}\}$ and $\{x_{n,k}\}$, where $k=1\ldots K$. $K$ is the number of keypoints. We provide two ways to apply a smooth the jump cut between $I_m$ and $I_n$ for a seamless transition. 

The first way is to add a $T$ number of intermediate frames between these two frames. The generation of each middle frame $I_t$ is guided by the linearly interpolated DensePose keypoints $\{(1-\alpha_t)x_{m, k} + \alpha_tx_{n, k}\}$, with $\alpha_t = t/(T+1)$. However, accompanying new frames with silent audio results in an awkward, broken speech in some cases. For example, when a person is speaking quickly, but with short  ``umms'' or ``uhs'' in the middle, if we remove these filler words and fill in with new frames with silence, the resulting video will sound inarticulate.

\begin{figure}[htbp]
    \centering
    \includegraphics[width=.9\linewidth]{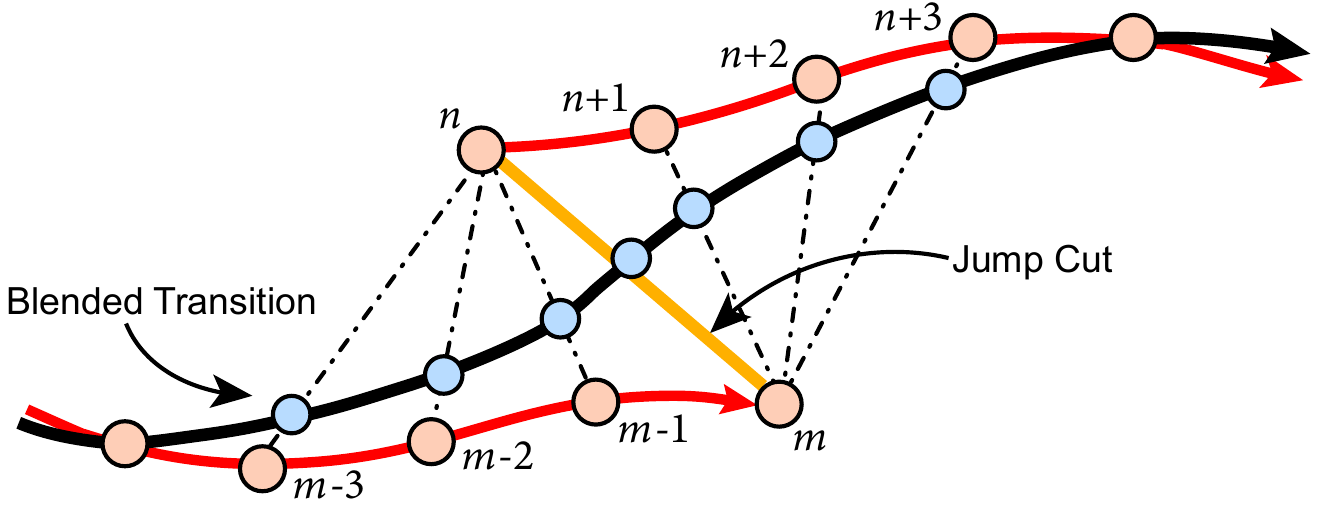} 
    \caption{{\bf Blended transitions.} We modify the frames before and after the jump cut, blending them smoothly. This allows us to smooth the jump cut without inserting additional frames.}\label{fig:blended}
    \vspace{-0.3cm}
\end{figure}

Thus, we provide another method, called {\it blended transition}, in order to avoid this audio artifact. Similarly to Zhou et al.~\cite{zhou2022audio}, we synthesize blended frames to replace original frames around a small temporal neighborhood with a synthetic transition, so that the video can
smoothly transit from frames before $I_m$ to frames after $I_n$, as shown in Fig.~\ref{fig:blended}.

We define the neighborhood using the frame range $[m-H, m]$ and
$[n, n + H]$, with $H$ as neighborhood size. We choose $H = 4$ in our experiments. Each frame $I_i$ is blended with frame $I_n$, where $i \in [m-H, m]$, with weight $\alpha_i\in [0, \tfrac{1}{2H}, \ldots, \tfrac{1}{2}]$, with DensePose keypoints blended by $\{x'_{i, k}=(1-\alpha_i)x_{i, k} + \alpha_ix_{n, k}\}$. Similarly, each frame $I_j, k \in [n, n+H]$ is blended with frame $I_m$ with weight $\alpha_j\in[\tfrac{1}{2}, \tfrac{H+1}{2H},\ldots, 1]$ and the corresponding DensePose keypoints are  $\{x'_{j, k}=(1-\alpha_k)x_{j, k} + \alpha_jx_{m, k}\}$. The blended frames are resynthesized and guided by the blended, dense keypoints. The blended transition does not change the existing number of frames in the video and thus does not suffer the audio insertion issue. Please refer to the supplementary for the video demo with audio for the filler words removal jump cut smoothing.

\section{Experiments}
\begin{figure}[h!]
    \centering
    \setlength{\tabcolsep}{1pt}
    \begin{tabular}{ccc}
    \includegraphics[width=0.15\textwidth]{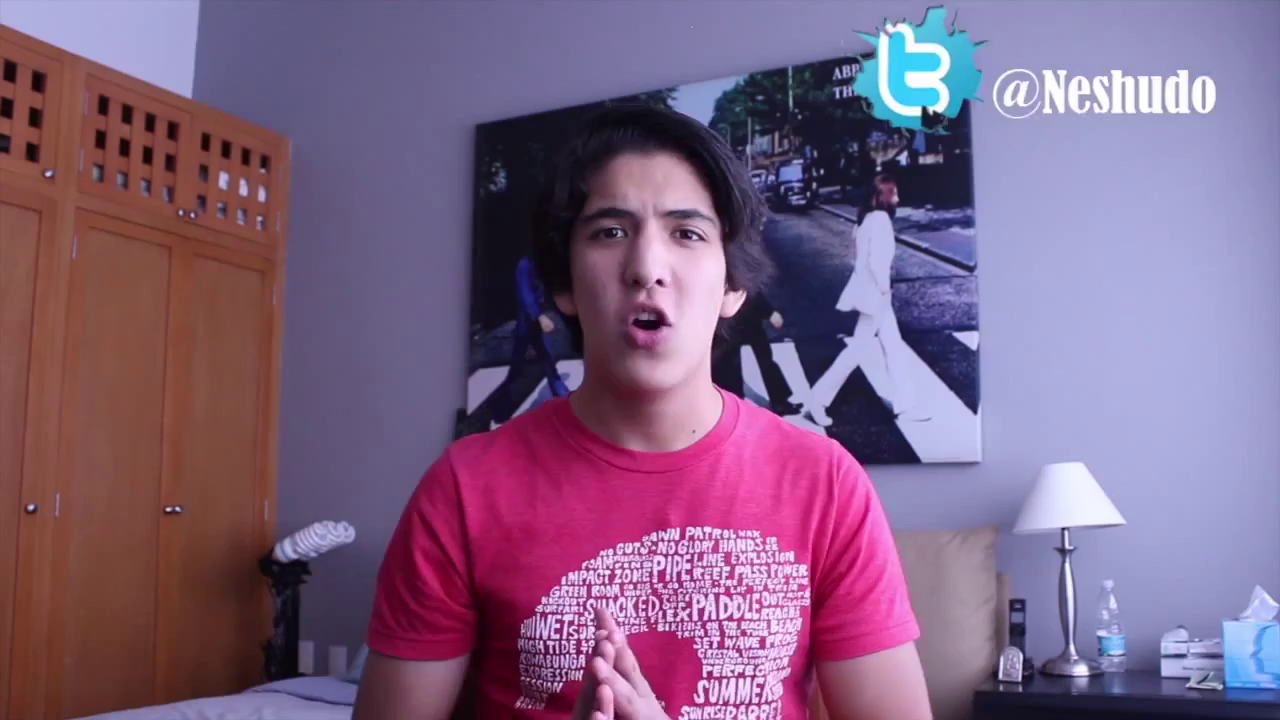} &\includegraphics[width=0.15\textwidth]{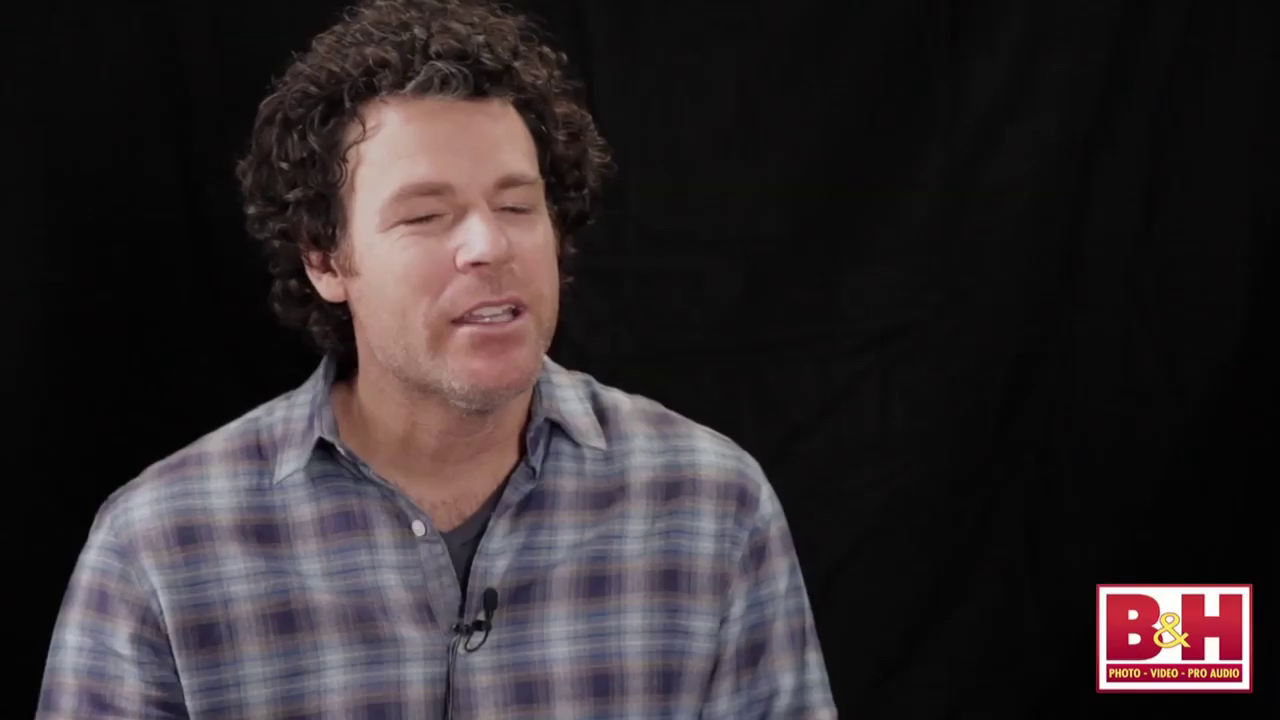}
    &\includegraphics[width=0.15\textwidth]{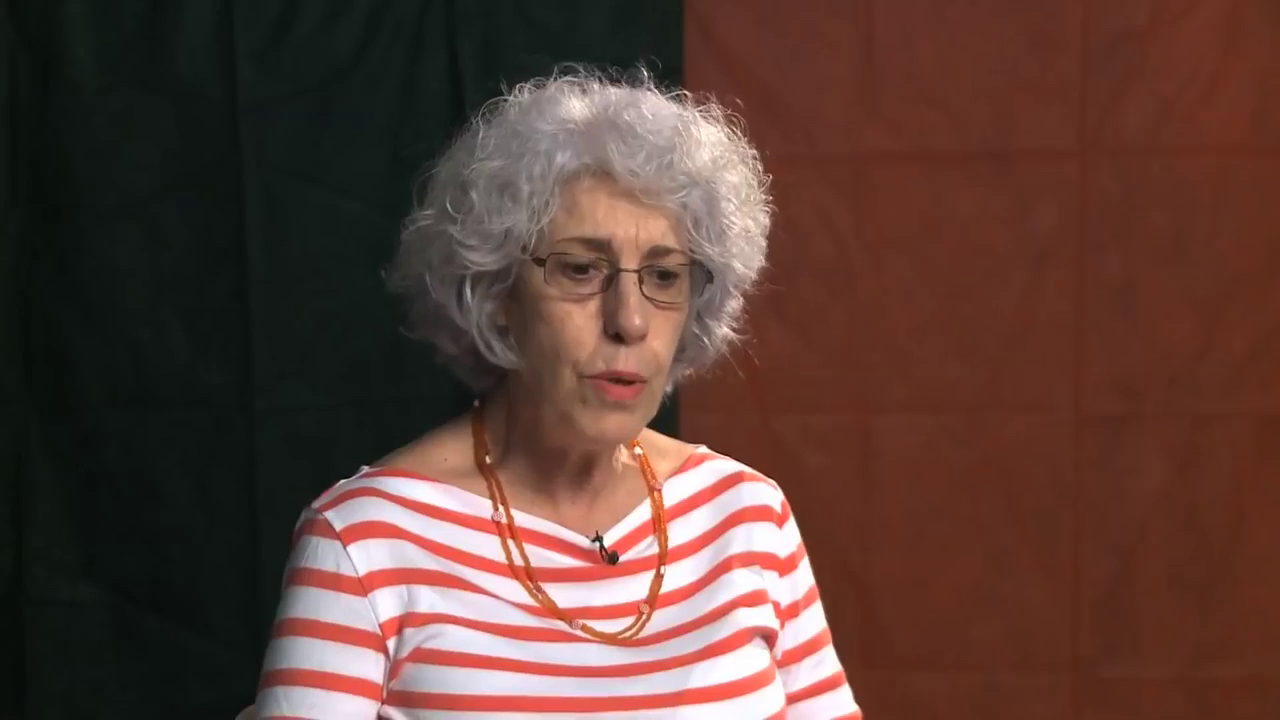} \\
    \includegraphics[width=0.15\textwidth]{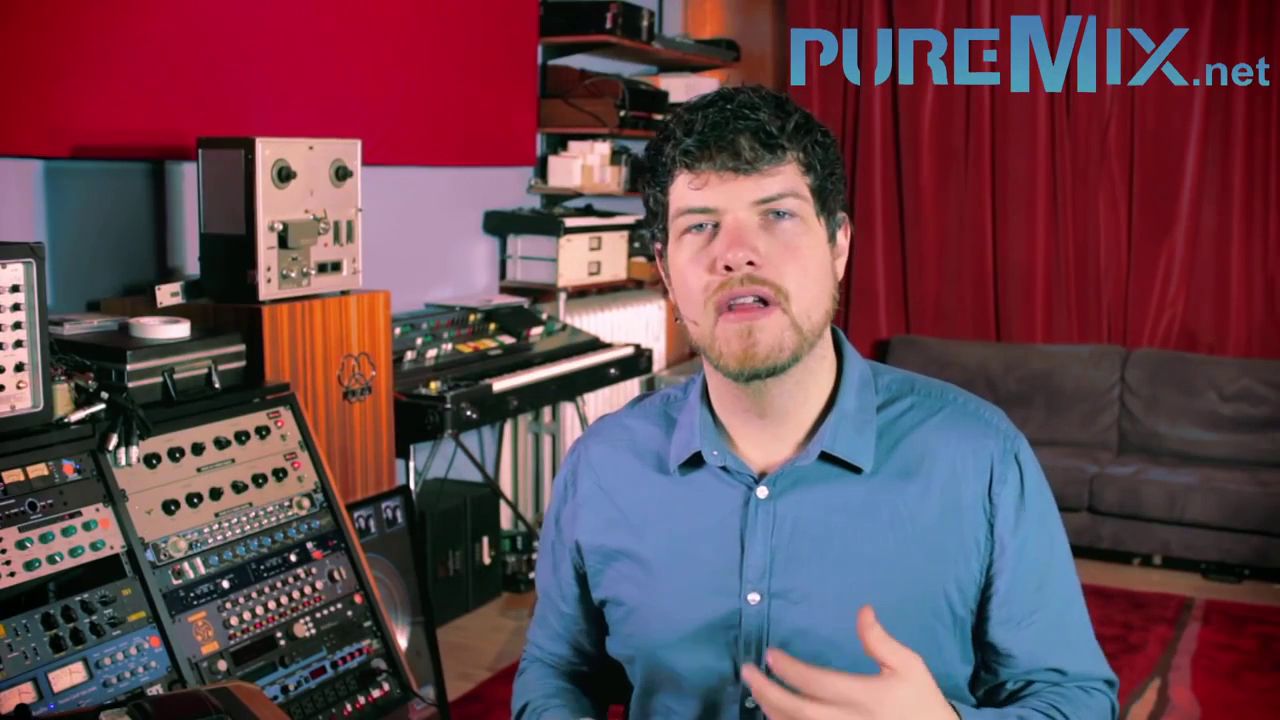} &\includegraphics[width=0.15\textwidth]{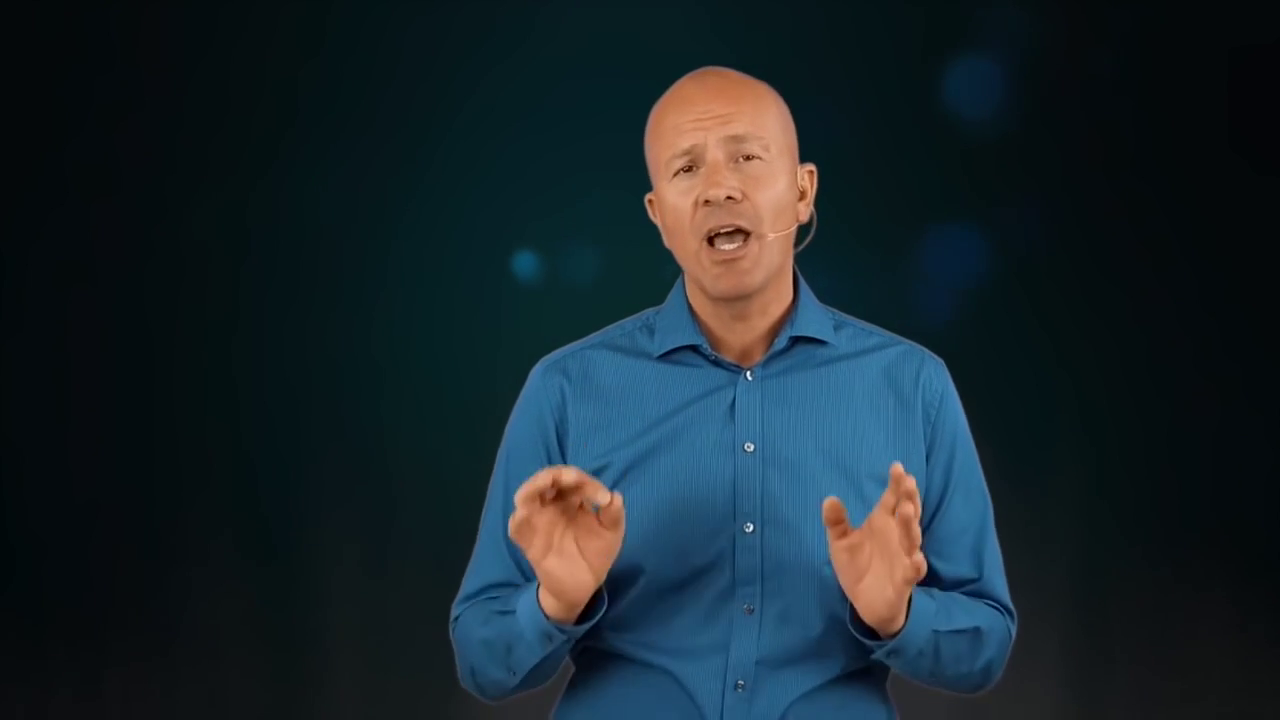}
    &\includegraphics[width=0.15\textwidth]{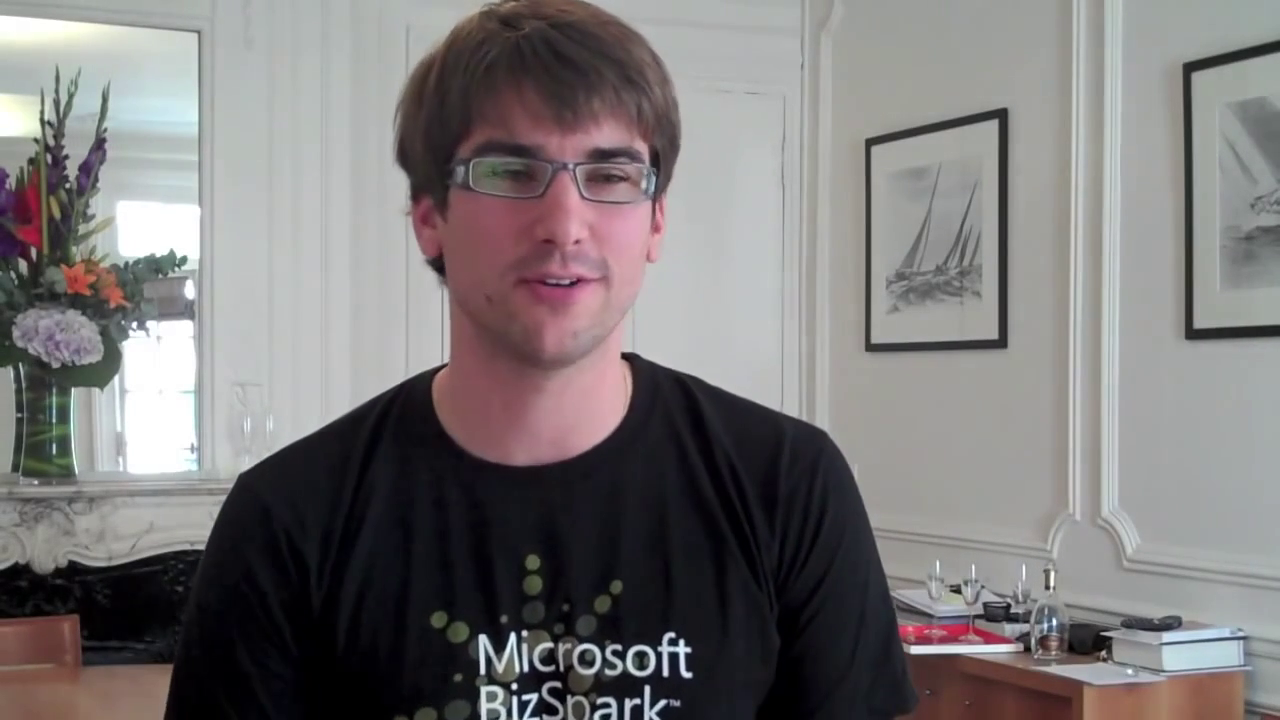} \\
    \end{tabular}
    \caption{\textbf{Examples of our talking head videos.} We collect 600 talking head video clips and smooth jump cuts corresponding to filler words.}\label{fig:dataset_example}
    \vspace{-0.2cm}
\end{figure}

\begin{figure}[h!]
    \centering
    \def\imW{0.09\textwidth}
    \setlength{\tabcolsep}{1pt}
    \begin{tabular}{c|ccc|c}
    \includegraphics[width=\imW]{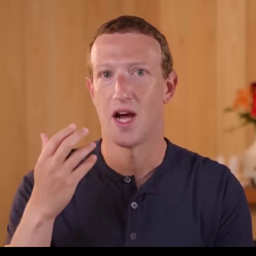} 
    &\includegraphics[width=\imW]{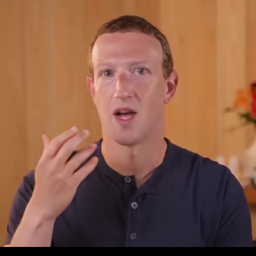}
    &\includegraphics[width=\imW]{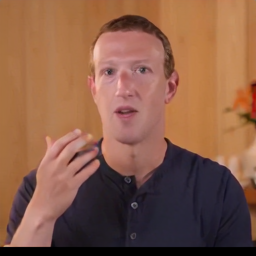}
    &\includegraphics[width=\imW]{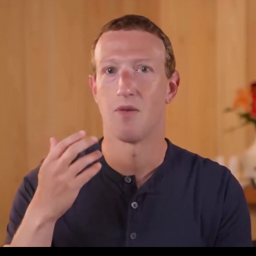}
    &\includegraphics[width=\imW]{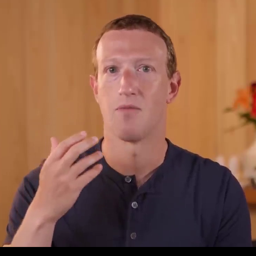}
    \\
    \includegraphics[width=\imW]{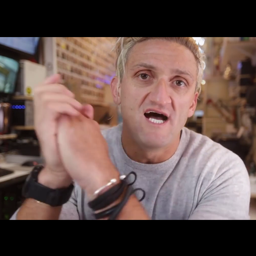} 
    &\includegraphics[width=\imW]{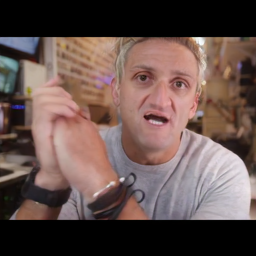}
    &\includegraphics[width=\imW]{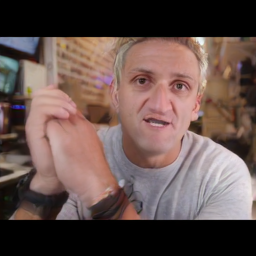}
    &\includegraphics[width=\imW]{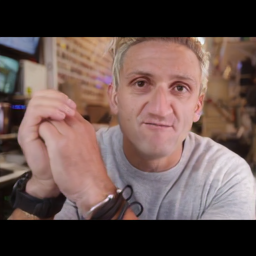}
    &\includegraphics[width=\imW]{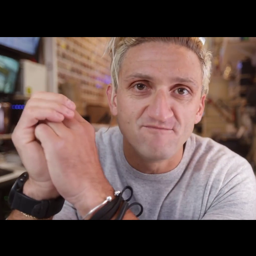}   
    \\
    \includegraphics[width=\imW]{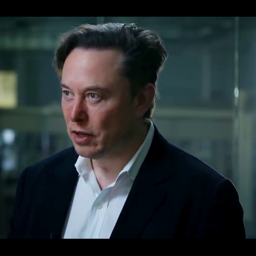} 
    &\includegraphics[width=\imW]{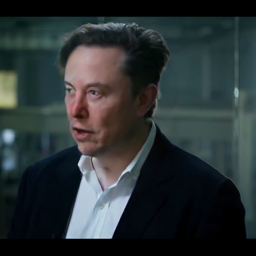}
    &\includegraphics[width=\imW]{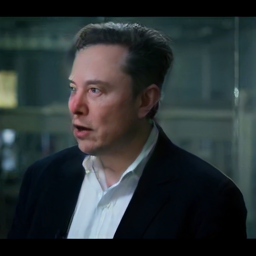}
    &\includegraphics[width=\imW]{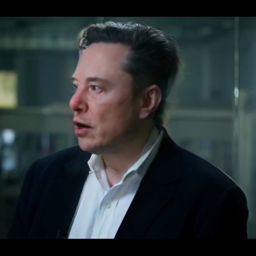}
    &\includegraphics[width=\imW]{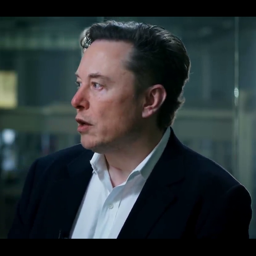}
    \\
    \includegraphics[width=\imW]{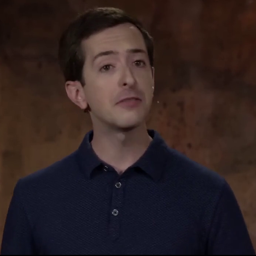} 
    &\includegraphics[width=\imW]{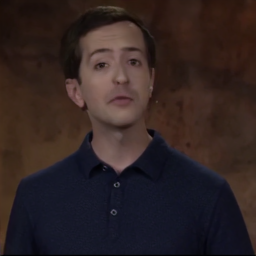}
    &\includegraphics[width=\imW]{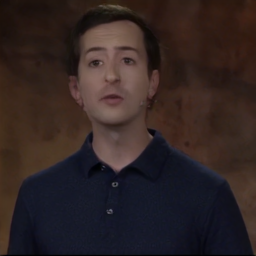}
    &\includegraphics[width=\imW]{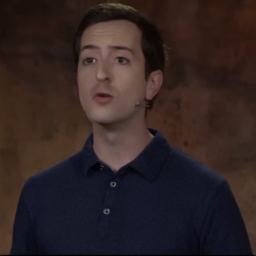}
    &\includegraphics[width=\imW]{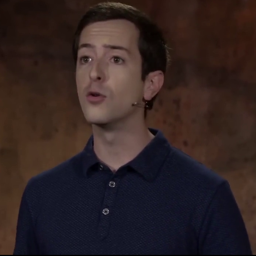}
    \\
    \includegraphics[width=\imW]{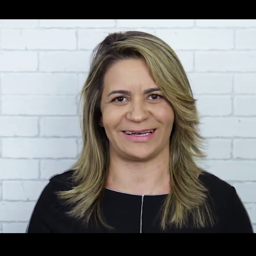} 
    &\includegraphics[width=\imW]{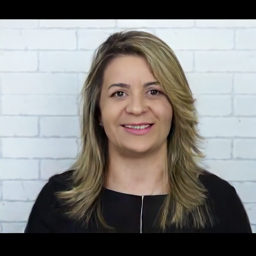}
    &\includegraphics[width=\imW]{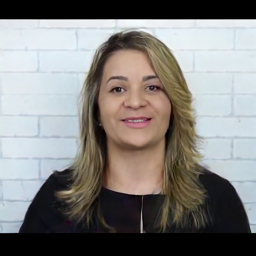}
    &\includegraphics[width=\imW]{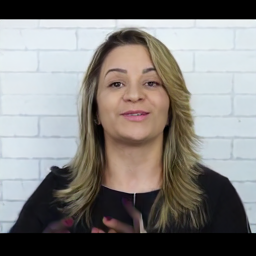}
    &\includegraphics[width=\imW]{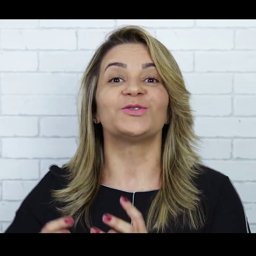}
    \end{tabular}
    \caption{{\bf Examples of our synthesized transition sequence} from the most left frame to the most right frame. Our method can create a seamless transition under different head pose changes. Row \#2: head moving from back to front; Row \#3: head rotating from front view to the side view; Row \#4: head rotating from one side to another side. Row \#5: head moving up and down.}\label{fig:sequence_vis}
\end{figure}

We show our synthesized transition sequence under diverse jump cut situations. See Fig.~\ref{fig:sequence_vis} for selected examples. Our method successfully achieves seamless transition under challenging head pose changes such as extreme rotation or the back-and-forth movement of the head. We show our dataset collection and pre-processing in Sec.~\ref{sec:data}, and the comparison with baselines in Sec.~\ref{sec:eval}. We further analyze how using more source frames from the video with our proposed attention mechanism improves synthesis quality in Sec.~\ref{sec:attention}. Finally, we demonstrate how we apply our jump cut smoothing technique for removing filler words in talking head videos.

\subsection{ Dataset and pre-processing.}\label{sec:data} We target our jump cut smoothing application for the talking videos that contain upper torso. Note that this is a wider (and hence more challenging) cropping scheme than many existing talking head video datasets that only contain facial portions. Therefore, we collected $600$ 720p video clips from the raw AVSpeechDataset collection~\cite{ephrat2018looking} for training and $50$ videos for testing. Each video contains a person talking either facing towards the camera or the interviewer (see Fig.~\ref{fig:dataset_example}) for $10$ to $20$ seconds in a static background. While we do not add any manual annotation, we use Detectron2~\cite{wu2019detectron2} to detect the DensePose body part coordinate map per frame, and quantize the DensePose uv map to turn them into keypoints. For more accurate facial expression representation, we augment the DensePose keypoints with face landmarks using HRNet~\cite{sun2019deep}.  For training, we crop the person with the detected DensePose bounding box, and resize to $256\times 256$.  In the training stage, we randomly pick two source and a target frames. At inference time, our method can be easily extended to incorporate more source images (Section~\ref{sec:attention}).

{\noindent \bf Comparison to image animation methods (N/A).}
\begin{figure}[!h]
    \centering
    \setlength{\tabcolsep}{0.2pt}
    \def\imW{0.09\textwidth}
    \begin{tabular}{ccccc}
    \includegraphics[width=\imW]{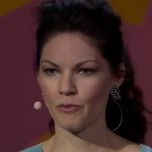} 
    &\includegraphics[width=\imW]{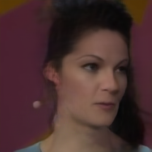}
    &\includegraphics[width=\imW]{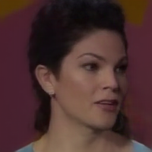}
    &\includegraphics[width=\imW]{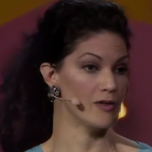}
    &\includegraphics[width=\imW]{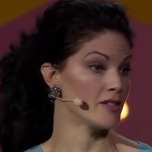} \\
     \includegraphics[width=\imW]{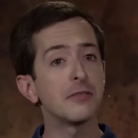} 
     &\includegraphics[width=\imW]{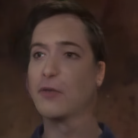}
     &\includegraphics[width=\imW]{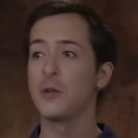}
    &\includegraphics[width=\imW]{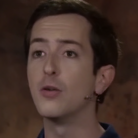}
    &\includegraphics[width=\imW]{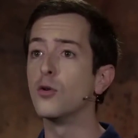} \\
     \includegraphics[width=\imW]{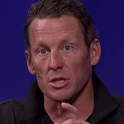} 
     &\includegraphics[width=\imW]{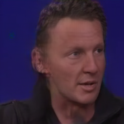}
     &\includegraphics[width=\imW]{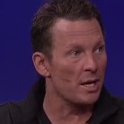}
    &\includegraphics[width=\imW]{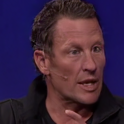}
    &\includegraphics[width=\imW]{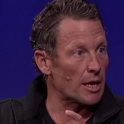} \\
    \makecell{\small Cut frame 1} &\makecell{\small  Face2Face$^{\rho}$} &\makecell{\small Face-Vid2Vid}& \makecell{\small Ours} & \makecell{\small Cut frame 2}
    \end{tabular}
    \caption{Comparison of synthesized images of Face2Face$^{\rho}$, and Face-Vid2Vid and our method.  Face2Face$^{\rho}$ generates distorted images, while Face-Vid2Vid loses sharpness and facial identity: microphones are gone and hair color does not match cut frame 2.}\label{fig:fvid2vid}
    \vspace{-0.5cm}
\end{figure}

\begin{figure*}[htbp]
    \centering
    \setlength{\tabcolsep}{1pt}
    \def\imW{0.2\textwidth}
    \begin{tabular}{ccccc}
    \includegraphics[width=\imW]{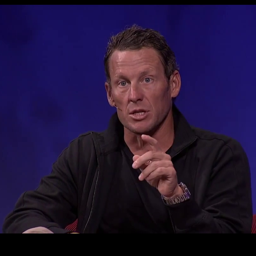}
    &\includegraphics[width=\imW]{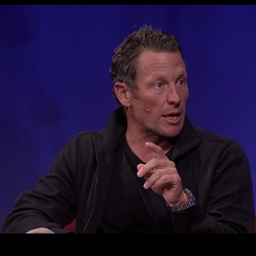}
    &\includegraphics[width=\imW]{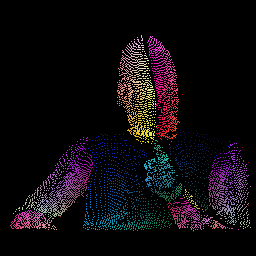} 
     &\includegraphics[width=\imW]{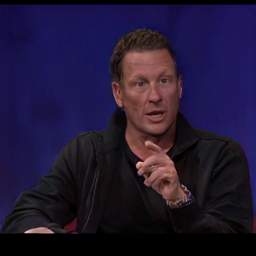} 
     &\includegraphics[width=\imW]{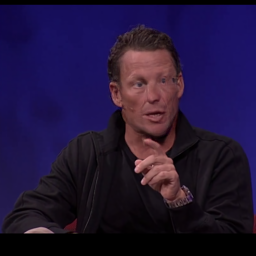} \\
      \includegraphics[width=\imW]{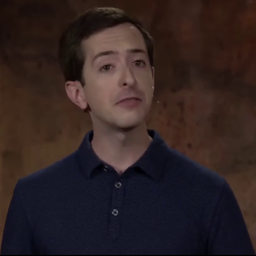}
    &\includegraphics[width=\imW]{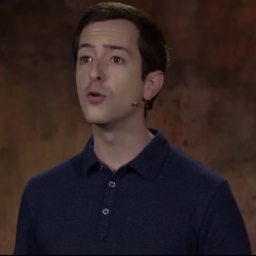}
    &\includegraphics[width=\imW]{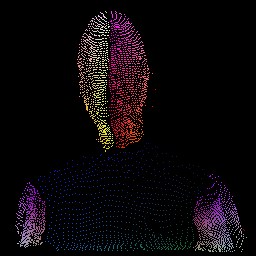} 
     &\includegraphics[width=\imW]{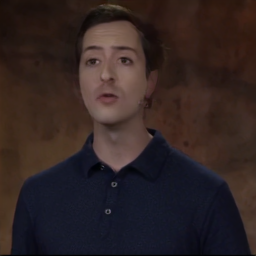} 
     &\includegraphics[width=\imW]{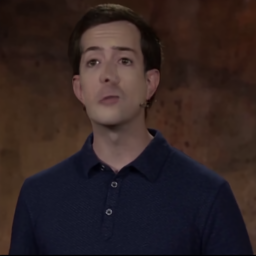}\\
     \includegraphics[width=\imW]{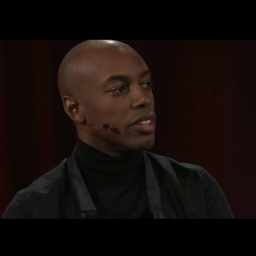}
     &\includegraphics[width=\imW]{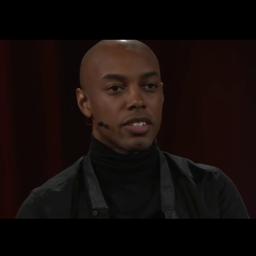}
    &\includegraphics[width=\imW]{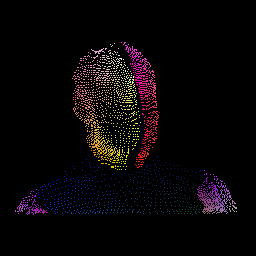} 
     &\includegraphics[width=\imW]{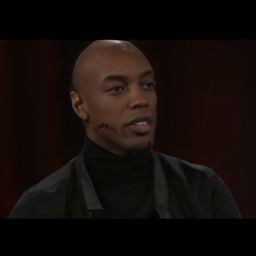} 
     &\includegraphics[width=\imW]{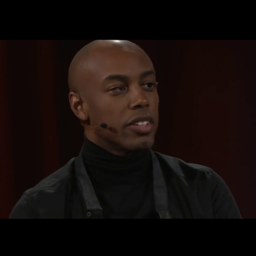}\\
     \includegraphics[width=\imW]{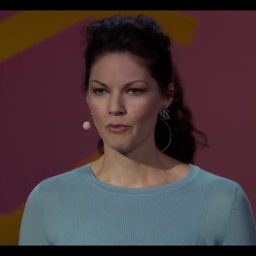}
     &\includegraphics[width=\imW]{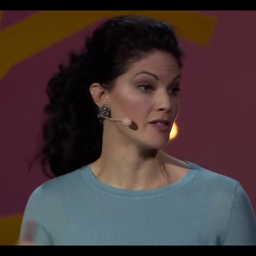}
    &\includegraphics[width=\imW]{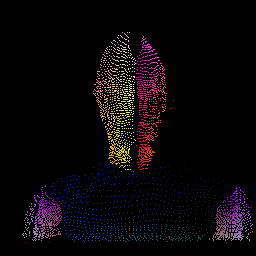} 
     &\includegraphics[width=\imW]{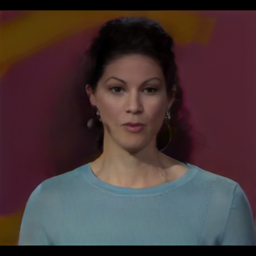} 
     &\includegraphics[width=\imW]{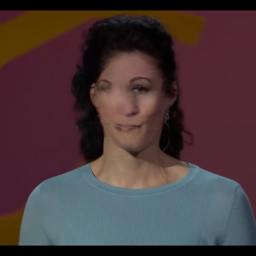}\\
     \small{a) Jump cut frame 1} &\small{b) Jump cut frame 2} & \small{c) Our kpts} & \small{d) Our results} & \small{e) FILM results} 
    \end{tabular}
    \caption{{\bf Qualitative comparison with FILM} for selected synthesized transition frame from jump cut end frame 1 to frame 2.}\label{fig:filmcomp}
\end{figure*}

\begin{table*}[htbp]
    \centering
    \setlength{\tabcolsep}{1pt}
    \begin{tabular}{ cccccccccc }
    \toprule
    \multicolumn{1}{c}{} 
     &\multicolumn{1}{c}{All}
     & \multicolumn{2}{c}{rotation $\ge$ $15^{\circ}$}
        & \multicolumn{2}{c}{rotation $\ge$ $30 ^{\circ}$}
        & \multicolumn{2}{c}{rotation $\ge$ $45 ^{\circ}$}
        & \multicolumn{2}{c}{rotation $\ge$ $60 ^{\circ}$} \\

        \cmidrule(lr){2-2} \cmidrule(lr){3-4} \cmidrule(lr){5-6} \cmidrule(lr){7-8} \cmidrule(lr){9-10}
    Methods  &FID~\cite{heusel2017gans}$\downarrow$  &PSNR$\uparrow$ &ArcFace~\cite{deng2019arcface}$\uparrow$  &PSNR$\uparrow$ &ArcFace~\cite{deng2019arcface}$\uparrow$  &PSNR$\uparrow$ &ArcFace~\cite{deng2019arcface}$\uparrow$  &PSNR$\uparrow$ &ArcFace~\cite{deng2019arcface}$\uparrow$ \\
    \midrule
    FILM              &4.99  &25.37 &0.51  &25.68 &0.39  &25.65 &0.33 &25.27 &0.23 \\
    Ours. + FILM kpts   &{\bf 4.35} &25.29 &0.51  &25.57 &0.41 &25.55  &0.35 &25.23  &0.25 \\
    Ours.  &{\bf 4.53}  &25.19 &{\bf 0.52} &25.43 &{\bf 0.45} &25.38 &{\bf 0.40}  &24.84 &{\bf 0.31} \\
 \bottomrule
\end{tabular}
    \caption{{\bf Quantitative comparison with FILM} in terms of synthesized frame realism and reconstruction fidelity. Here ``rotation" means the head rotation along the yaw axis.` `Ours+FILM kpts" means synthesize the image using DensePose keypoints interpolated with FILM. Our method outperforms FILM in terms of FID and ArcFace similarity, especially in the cases where head rotation happens in the jump cut.}
    \label{tab:quan}
    \vspace{-0.4cm}
\end{table*}

\subsection{Evaluation}\label{sec:eval} 
We show how image animation techniques {\it cannot} be applied to our jump cut smoothing problem in Fig.~\ref{fig:comp_to_image_animation}. More specifically, FOMM~\cite{siarohin2019first} and ImplicitWarping~\cite{mallya2022implicit} cannot generate new transition frames \textit{without} driving videos, as their latent motion representations lack 3D correspondence. The same key point could correspond to different head regions (see Row\#2 in Fig.~\ref{fig:comp_to_image_animation}). Face-Vid2Vid~\cite{wang2021one} and Face2Face$^{\rho}$ animate only one image,
necessitate using just one of the end frames as the source. Fig.~\ref{fig:fvid2vid} presents the  generated frame  just before the second jump cut end frame. Evidently, the resulting images sacrifice facial identity and clarity, leading to another abrupt jump in the transition.

\vspace{0.1cm}
{\noindent \bf Comparison to frame interpolation methods.} We compare our method for jump cut smoothing against the state-of-the-art frame interpolation method FILM~\cite{reda2022film}. For testing, we randomly pick an equally spaced triplet, and run each method to synthesize the middle frame based on the two end frames. For our method, the keypoints of the middle frame are generated by linearly interpolating the DensePose keypoints of the two end frames. Furthermore, we also report a variant that runs FILM in the keypoint space, and then decodes the keypoints into pixels using our attention-based synthesis network. In Table~\ref{tab:quan}, 
we measure the realism of the synthesized frames against the target frames with Frechet Inception Distance (FID)~\cite{heusel2017gans}, using around $8000$ test frames. In addition, we measure the fidelity of synthesized image against the ground truth test frame with PSNR. We also measure how much the facial identity is preserved in the synthesized frames by reporting the ArcFace~\cite{deng2019arcface} cosine similarity between the facial embeddings of source and output frames. To further quantify how our method and FILM perform under different amount of head pose changes, we categorize the jump cut end frame difference by the head rotation degrees along the yaw axis, and report the evaluation metrics in each category. As shown in Table~\ref{tab:quan}, our method achieves lower FID score overall, meaning our method attains higher photorealism than FILM. We also achieve higher ArcFace similarity across all rotation angles. In particular, our advantage grows larger at more extreme head pose changes (e.g., rotation $\ge 60^{\circ}$). FILM, an optical flow-based method, likely fails to find correspondence when the head rotates from the front view (or more extremely one side view) to another side view,  causing distortion in the synthesized face (see the rightmost column in Fig.~\ref{fig:filmcomp} for reference). On the contrary, our method does not rely on the appearance feature to find correspondence, as we build upon the DensePose priors with attention mechanism to find better correspondence. Lastly, ``Ours+FILM kpts" achieves the lowest FID score, likely because FILM does a better job in synthesizing complex objects like hands than linear interpolation. 

\subsection{Attention with more frames}\label{sec:attention} 
\begin{figure}[htbp]
    \centering
    \setlength{\tabcolsep}{1pt}
    \begin{tabular}{cc}
    \multicolumn{2}{c}{\includegraphics[width=1.\linewidth]{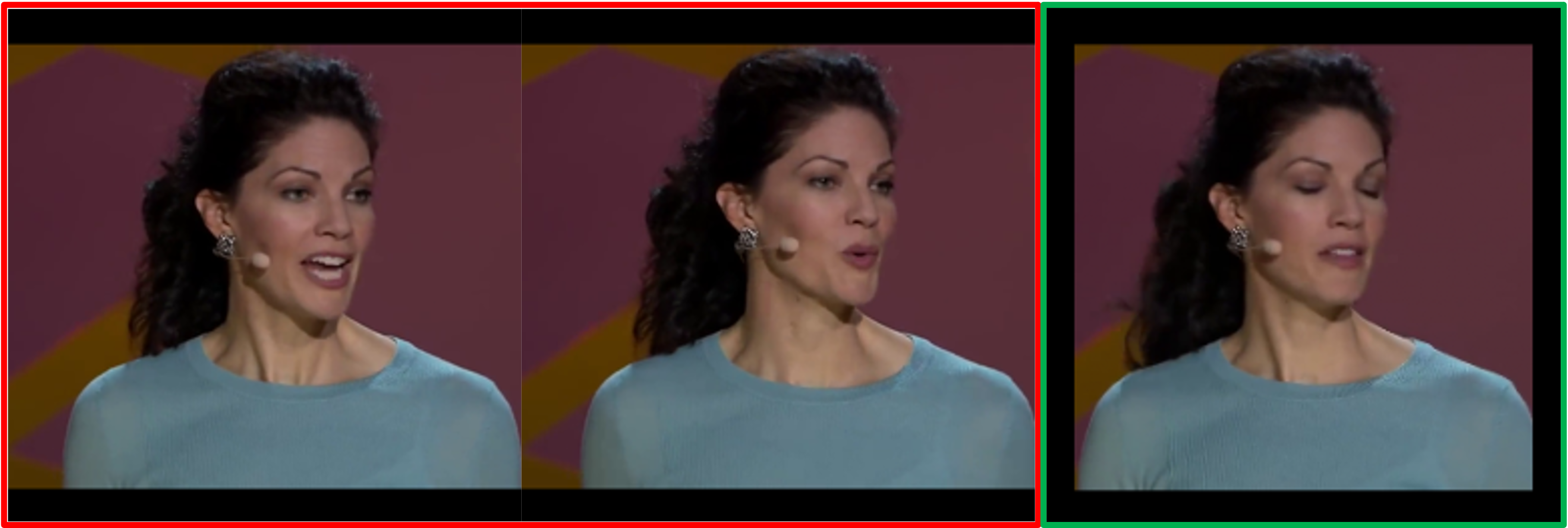}}\\
    \hline
    \includegraphics[width=0.5\linewidth]{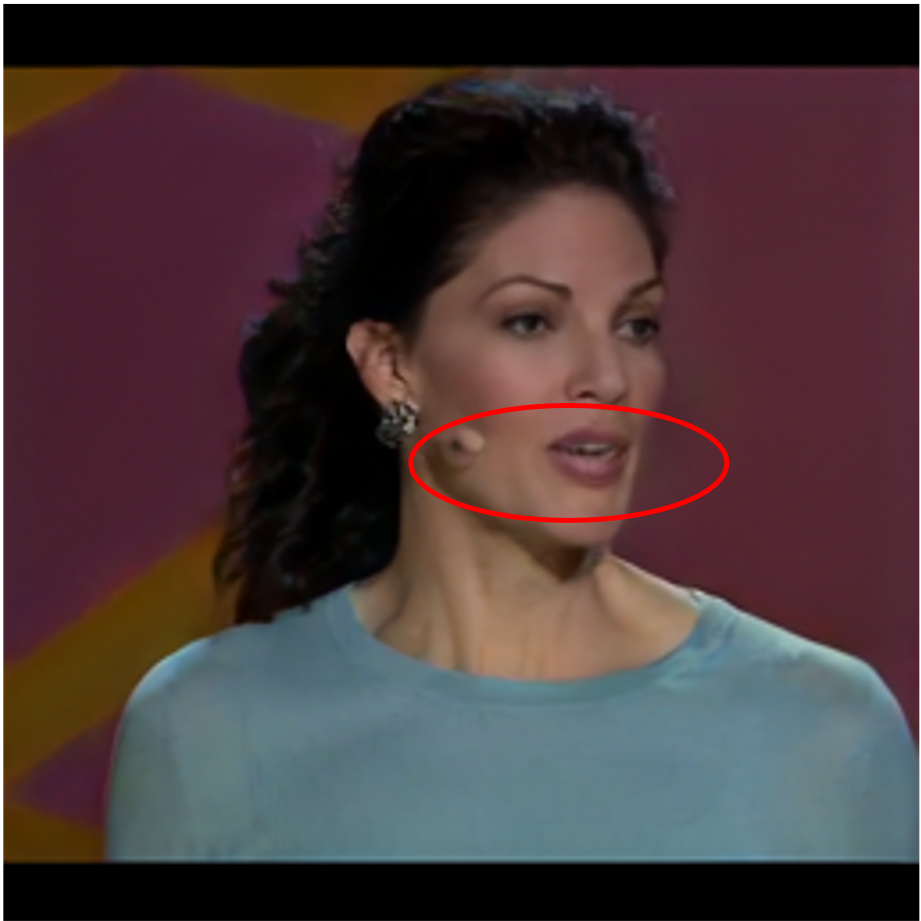} 
    &\includegraphics[width=0.5\linewidth]{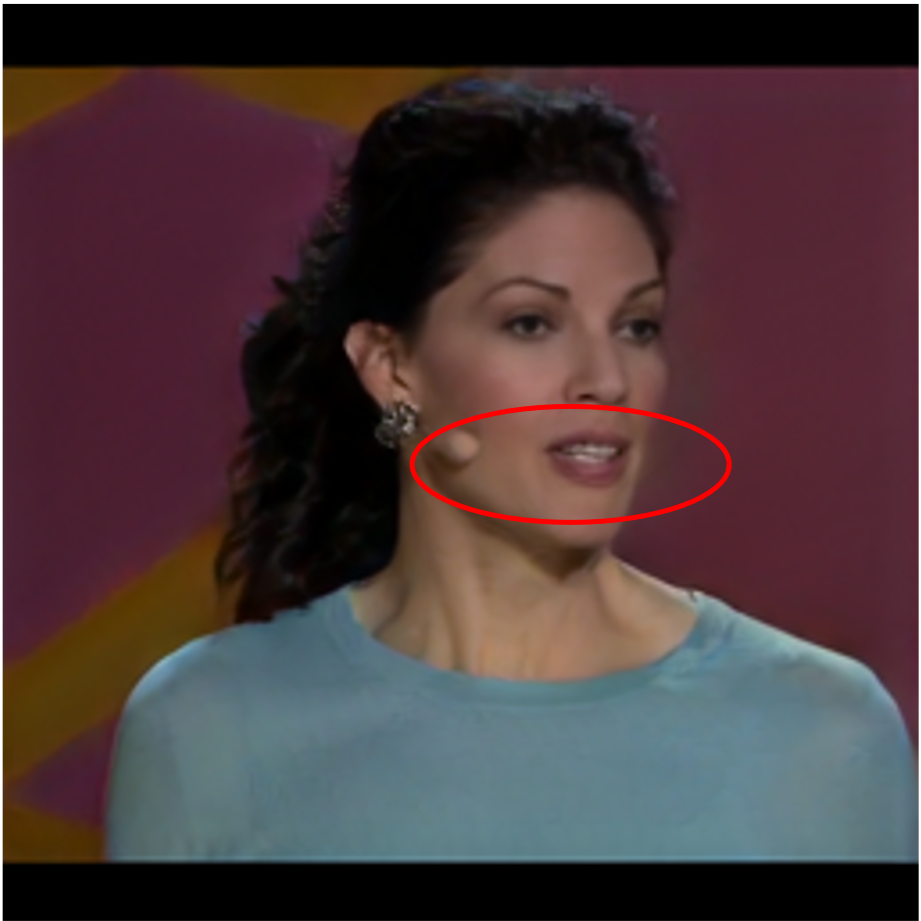} \\
    \end{tabular}
    \caption{{\bf Attention with more frames help improving the quality.} The top left two images within the red rectangle are the end frames of the jump cut. The top right image within green rectangle shows one of the 10 random images we add as additional sources to our attention. In the bottom row, we show our synthesized image using only the two end frames (left), and using the entire 12 source images (right). After adding $10$ extra randomly chosen frames from the video, our method generates more accurate teeth representation, thanks to the attention mechanism that can incorporate variable number of source frames as reference.}
    \label{fig:attn_artifacts}
    \vspace{-0.5cm}
\end{figure}

One big advantage of our method compared to flow-based frame interpolation methods is that at test time, we can accept a larger number of input source frames in the video to assist the synthesis process. With attention mechanism, our method can select the most appropriate feature per location among source frames. As shown in Fig.~\ref{fig:attn_vis}, different locations in the synthesize image correspond to their most relevant locations that are distributed among different sources. We find that this is especially helpful when certain parts are occluded in the two end frames, but are visible in the middle transition frames. For example, in Fig.~\ref{fig:attn_artifacts}, the speaker's mouth is widely open in one end frame and fully closed in the other end frame (see the images within red rectangle). If we only use these two end frames as source, the person's mouth in the generated middle frame looks unnatural, as there are no teeth features for a ``half-open" mouth in the end frames. However, there are teeth features among the other video frames. We add 10 extra frames randomly chosen from the video, and our method now generates a high-quality middle frame, with teeth exposed in the half-opened mouth.

\subsection{Jump cut smoothing for filler words removal} 
We further demonstrate our method on filler words removal video editing. We collected several talking videos where the person stutters and then apply filler words detection algorithm ~\cite{zhu2022filler} to cut out the filler words, resulting in unnatural jump cuts. We also manually remove some unwanted pauses and repetitive words to output a fluent talking video. Then we apply our jump cut smoothing with blended transition in Sec.~\ref{sec:blendedtransition} to the video. We show these demo videos with audio in the Supplementary Materials.

\vspace{0.3cm}
\section{Discussions and Limitations}
Our method can create smooth transition under diverse jump cut cases, especially for the head movement. However, we failed when there are complex hand gesture move. Synthesizing realistic hand is even more challenging because 1.) The video frames with hand movement often have motion blurs, which makes our network hard to discriminate real or fake hands. 2.) DensePose itself cannot model fine-grained hand features such as fingers. 3.)
The hands motion is more complicated than head. For example, when the speaker's hand moves from clenched to stretched, or from palm facing to back facing towards the camera, this non-planer motion  cannot be modeled with linearly interpolated key points.

\clearpage
{\small
\bibliographystyle{ieee_fullname}
\bibliography{main}
}

\section{Demo videos}
\begin{figure*}[htbp!]
    \centering
    \setlength{\tabcolsep}{0.5 pt}
    \begin{tabular}{ccccc}
    \includegraphics[width=0.2\textwidth]{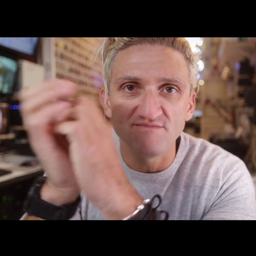} 
    &\includegraphics[width=0.2\textwidth]{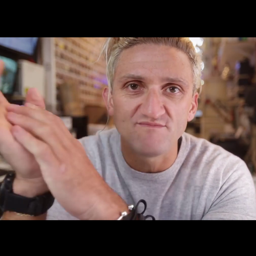} 
    &\includegraphics[width=0.2\textwidth]{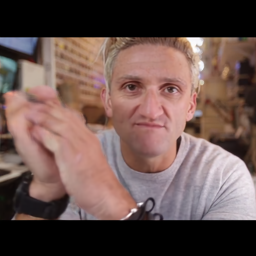}
    &\includegraphics[width=0.2\textwidth]{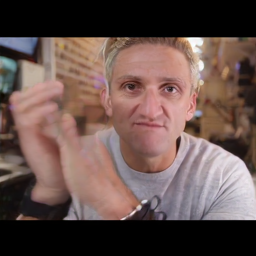}
    &\includegraphics[width=0.2\textwidth]{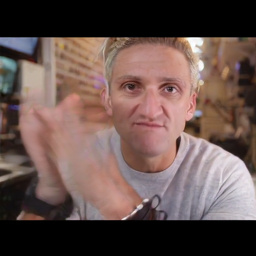}\\
    frame 1 & frame 2 & FILM  &Ours. + FILM kpts & Ours. \\
    \end{tabular}
    \caption{\textbf{Failure cases of hands synthesis.} When the speaker's hands gesture moves from frame 1 to frame 2, we show the synthesized middle frame with FILM, Ours. + FILM kpts, and Ours.}\label{fig:failure}
\end{figure*}
\begin{figure*}[htbp!]
    \centering
    \setlength{\tabcolsep}{0.5 pt}
    \begin{tabular}{ccccc}
    \includegraphics[width=0.2\textwidth]{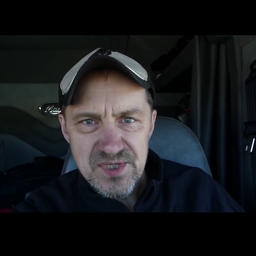} 
    &\includegraphics[width=0.2\textwidth]{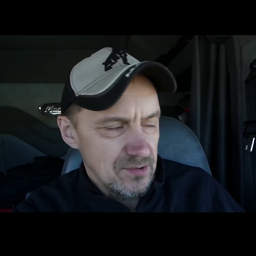} 
    &\includegraphics[width=0.2\textwidth]{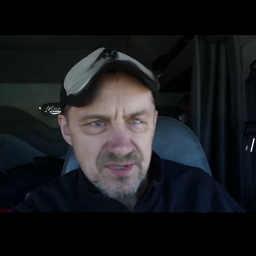}
    &\includegraphics[width=0.2\textwidth]{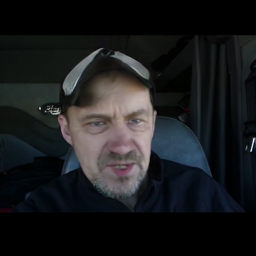}
    &\includegraphics[width=0.2\textwidth]{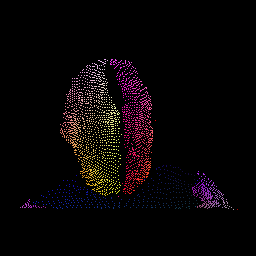}\\
    frame 1 & frame 2 & FILM  &Ours. & Ours. kpts \\
    \end{tabular}
    \caption{\textbf{Limitations of using DensePose representation.} When the speaker wears a hat and rotates from front view to side view,  we show the synthesized middle frame with FILM and Ours. The rightmost column shows the corresponding interpolated DensePose keypoints with our method. }\label{fig:failure2}
\end{figure*}

Please refer to our web page to see 1.) video demos (with audio) of applying our method on filler words removal video editing (Sec. 4.4 in the main paper), 2.) video results comparison between our method and FILM for jump cut smoothing,  3.) how we can control the transition sequence by face landmark manipulation, 4.) visualization of the learned correspondence via our attention mechanism.

\section{Network architecture}
Given multiple source images with the their extract DensePose keypoints augmented with face landmarks, and the target DensePose keypoints with face landmarks, our method firstly learns to use cross model attention to warp the source image features, then the warped image feature is concatenated with the jump cut end image features together to feed into the generator network to produce a target image (see. Fig. 2 in the main paper). We have  {\bf three} StyleGAN2 based {\bf encoders}: 1) $E_{V}$ to encode source image features served as {\it value}, 2)  $E_K$ to encode source dense keypoints features as {\it key}, 3)  $E_Q$ to encode target dense keypoints features as {\it query}. 

The network structures of these encoders are similar except for the input layer (see Tab.~\ref{tab:encoder}). For the target dense keypoints encoder $E_Q$, we concatenate the 3-channel DensePose keypoints (IUV) input with the Gaussian blurred 68-channel facial landmark one encoding map, producing a 71-channel input. For $E_K$, we additional concatenate the source RGB image with the dense keypoints, producing a 74-channel input. For image encoder $E_V$, the input is a 3-channel RGB image. 

\begin{table}[htbp]
    \centering
    \begin{tabular}{c|c}
    \hline
        256 $\times$ 256   &Input (with channels $C_{in}$)  \\ \hline
        256 $\times$ 256   &ConvLayer ($C_{in}\rightarrow 32$) \\ \hline
        128 $\times$ 128   &ResBlock down ($32\rightarrow 64$) \\ \hline
        64  $\times$ 64   &ResBlock down ($64\rightarrow 128$) \\ \hline
        64  $\times$ 64   &ConvLayer ($128\rightarrow 128$) \\ \hline
        64  $\times$ 64   &ProjectionLayer ($128\rightarrow 64$) \\ \hline
    \end{tabular}
    \caption{Encoder network structure.}
    \label{tab:encoder}
\end{table}
For the image generator part, we adapt the network structure in Co-Mod GAN to accept our warped feature at $1/4$ image resolution as input and generate the output image.The generator is a UNet network with skip connection between the encoder and decoder layers. Here our encoder and decoder are not symmetric since the input is the warped image feature, we only keep skip connections on the symmetric layers in the encoder and decoder. 

\section{Training details}
We train our three encoders, Co-Mod GAN generator and discriminator end-to-end with in the same way as  the original Co-Mod GAN did. Our models are trained on one Nvidia A40 GPU with a batch size of $8$, we use Adam optimizer with an initial learning rate of $0.0002$, and $(\beta_1, \beta_2)=(0, 0.99)$. For the DensePose keypoints discretization (Sec 3.1 in the main paper), we use $n=64$ for quantizing the UV map.

We adopt two-stage training, in the first stage, we sample source and target triplets from the video per iteration, with the corresponding extracted DensePose keypoints and facial landmarks from the video frame. In the second stage, to simulate the incomplete target DensePose keypoints due to linear interpolation in the inference time, we finetune the trained model in the first stage by only use visible DensePose in all source images for the target, and make the generator to learn to inpaint the hole and generate realistic image. 

\section{Discussions}
{\noindent \bf Hand gesture movement during the jump cut.}
Our method can create smooth transition under diverse jump cut cases, especially for the head movement. However, we failed when there are complex hand gesture move. For example, in Fig.~\ref{fig:failure}, when the speaker's hands moves from closed to stretched, this nonlinear motion cannot be modeled with linear interpolation on the hands dense keypoints, while FILM synthesizes slightly better than our method because  it learns flow based correspondence in the raw RGB pixel space, and thus our method with FILM interpolated DensePose keypoints produces more realistic image compared with linearly interpolated keypoints.

\vspace{0.2cm}
{\noindent \bf Limitations of DensePose representation.} We use pre-defined DensePose keypoints and facial landmarks extracted from the image itself as input motion. DensePose keypoints map the pixels in the image to points in the 3D human surface, and thus every keypoint has the 3D semantic correspondence, which enables us to do keypoints interpolation for the  transition sequence for the jump cut. In some cases, this keypoint correspondence fails with view point changing when the person wears some accessories.
For example, in Fig.~\ref{fig:failure2}, the person wears a hat, and part of the hat is overlapped with the forehead, and the thus forehead appearance is not consistent with different views because of the hat. Therefore, our synthesized image has slightly blurred artifacts in the forehead compared with FILM.

\end{document}